# scientific reports

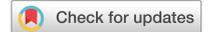

OPEN

# A computational framework for physics-informed symbolic regression with straightforward integration of domain knowledge

Liron Simon Keren[1]✉, Alex Liberzon[1] & Teddy Lazebnik[2]

Discovering a meaningful symbolic expression that explains experimental data is a fundamental challenge in many scientific fields. We present a novel, open-source computational framework called *Scientist-Machine Equation Detector* (SciMED), which integrates scientific discipline wisdom in a scientist-in-the-loop approach, with state-of-the-art symbolic regression (SR) methods. SciMED combines a wrapper selection method, that is based on a genetic algorithm, with automatic machine learning and two levels of SR methods. We test SciMED on five configurations of a settling sphere, with and without aerodynamic non-linear drag force, and with excessive noise in the measurements. We show that SciMED is sufficiently robust to discover the correct physically meaningful symbolic expressions from the data, and demonstrate how the integration of domain knowledge enhances its performance. Our results indicate better performance on these tasks than the state-of-the-art SR software packages , even in cases where no knowledge is integrated. Moreover, we demonstrate how SciMED can alert the user about possible missing features, unlike the majority of current SR systems.

Modern research is constructed from three main phases: observation, hypothesis generation, and hypothesis validation[1–3]. During the observation phase, researchers collect data about the world, which later, during the hypothesis generation phase, is used to generate a hypothesis that explains this data. A good explanation commonly allows for extrapolation and, thus, the prediction of new data of the same observed system during the hypothesis validation phase[4,5]. A common way of hypothesis generation is Symbolic Regression (SR), where researchers discover a symbolic expression (sometimes noted as an equation or a symbolic function) that accurately matches a given dataset[6–8]. To be exact, researchers assume a set of measurements or features are taking part in some natural phenomena and record samples of these measurements. Intuitively, the SR task is to unveil a symbolic expression for the function connecting the experimental measurements[9].

SR stands at the root of many fields of research such as engineering[10], psychology[11], economy[12], physics[13], chemistry[14], and others[15] since all mathematically expressed models are formally a function. Thus, the hypothesis generation process in all of these fields can be viewed as the discovery of a function that allows us to determine a value of interest, given a set of related measurements. As a result, multiple computational frameworks have been proposed to automate this task[16]. These frameworks of SR require finding the optimal model structure, input parameters, and the algebraic functions connecting them at once, as opposed to linear regression, for example, that operates under the assumption that the source measurements and the target measurement are linearly dependent[17]. Though this somewhat simplistic assumption produced many useful models[18–20] via simple computations of a system of linear equations, it does not work for non-linear cases, which seem to dominate most (if not all) fields of science[21–23]. The general symbolic regression problem remains unsolved and super-exponential to the number of measurements, making it infeasible to brute-force for even medium-sized datasets. Indeed, SR is known to be an NP-hard task[24].

SR and other NP-hard tasks, such as *Vertex Cover*[25] and *Hamiltonian Circuit Completion*[26], are of great interest and hence multiple approximations to the optimal outcome of these tasks were investigated[27]. In particular, one can divide these attempts into two main groups: analytical-based and heuristic-based approximations. The first aims to find an algorithm that can approximate the optimal solution within some bounded error, in an asymptotic polynomial time, based on the data alone. The objective of the latter approach is to produce a working solution within a much smaller time frame than that of the prior. Practically, heuristic-based approaches add assumptions to the task to divide it into simpler cases that can be solved independently, and well-approximated cases that satisfy the assumptions. The heuristic-based approach has shown promising results in general and in SR in particular, using methods such as genetic algorithms[28], and sparse regression[29].

[1]Turbulence Structure Laboratory, School of Mechanical Engineering, Tel Aviv University, Tel Aviv, Israel. [2]Department of Cancer Biology, Cancer Institute, University College London, London, UK. ✉email: lirons.gb@gmail.com

Though these methods provide promising results, they do not consider valuable domain knowledge that their expert users (from now on referred to as scientists) can provide to help direct the regression efforts. To tackle this shortcoming, we present the Scientist-Machine Equation Detector (SciMED) system, designed to deduce equations using four levels of search and optimization methods, structured to direct more attention and resources to promising search directions, somewhat similar to the search route of a scientist. The novelty of the proposed work lies in the straightforward integration of domain knowledge, specific to the current SR task, by using several input junctions throughout the system's pipeline. Opposed to other physics-informed SR systems, this means that SciMED does not attempt to apply general rules for all physical SR tasks but instead allows the scientist to direct the search process with more precise information. This leads to more credible results and reduces the computational time and resources required by SciMED compared to other SR systems. It also aids in formulating an accurate symbolic expression even from data that contains high noise levels. Additionally, SciMED offers a novel a-priori feature selection process that enables scientists to test different hypotheses efficiently.

The rest of this paper is organized as follows. First, we review the current state-of-the-art SR systems. Afterwards, we formally introduce SciMED. Subsequently, we present five experiments representing the cases SciMED aims to tackle with their results. Lastly, we summarize our conclusions and discuss opportunities for future work.

## Related work

The task of fitting a numerical or symbolic function on a set of data points is common in multiple fields of research[30]. As opposed to a regression task, which provides a model structure and fits it to available data, symbolic regression (SR) simultaneously searches for a model and its parameters[31]. Due to the constantly increasing amount of data and computational capabilities, multiple attempts to automate data transformation into knowledge using SR have been proposed[32].

The process of automating SR faces multiple challenges, such as an exponentially sizeable combinatorial space of symbolic expressions leading to a slow convergence speed in many real-world applications[33] or increased sensitivity to overfitting stemming from unjustifiably long program length[34].

SR can be especially useful in physics[35], frequently dealing with multivariate noisy empirical data from non-linear systems with unknown laws[36]. Moreover, the SR's output must retain dimensional homogeneity, meaning all terms in SR expression have to carry the same dimensional units. On the one hand, this constraint reduces the potential search space for the SR, while on the other hand, it introduces a meta-data on the model construction that one needs to consider and handle[37,38] stated that symbolic equations in physics broaden human understanding by a) exposing expressions that can be indicators of underlying physical mechanisms, as well as b) identifying metavariables (variable combinations or transformations) that might ease later empirical modeling efforts. This sort of explainability helps to examine how the model's behavior, variables, and metavariables correspond to available prior knowledge in the field.

In this section, we provide an overview of the advantages and limitations of current SR methods. In addition, we focus on knowledge integration methods in the context of SR systems and review the state-of-the-art methods of SR that SciMED will be compared to in the experiments section.

### SR methods.
There are numerous methods for performing SR [37,39] that we can loosely divide into four main groups, depending on the underlying computational technique: brute-force search, sparse regression, deep learning, and genetic algorithms.

Brute-force search-based SR systems are, in principle, capable of successfully solving every SR task as they test out all possible equations to find the best performing one[40]. However, in practice, a naive implementation of brute-force methods is infeasible, even on small-sized datasets, because of its computational expense. Furthermore, these systems tend to overfit given large and noisy data[41], which is the case of typical empirical results in physics. Two main methods to overcome the computational expense are performed by[42,43], where they apply a brute-force approach on a reduced search space rather than perform an incomplete search in the entire search space. In both methods, the search space is reduced by removing algebraically equivalent expressions, either through the recursive application of the grammar production rules[42] or by preventing semantic duplicates using grammar restrictions and semantic hashing[43].

Sparse regression systems can substantially reduce the search space of all possible functions by identifying parsimonious models using sparsity-promoting optimization. A recognized sparse SR algorithm explicitly built for scientific use cases is proposed by[44] called *SINDy*. *SINDy* uses a Lasso linear model for sparse identification of non-linear dynamical systems that underlie time-series data. *SINDy*'s algorithm iterates between a partial least-squares fit and a thresholding (sparsity-promoting) step. Since its inception, *SINDy* has been continuously improved. For example,[45] increased its ability to solve real-time problems given noisy data,[46] added optimal model selection over various values of the threshold, and[47] have created *PySINDy*; an open-source Python package for applying *SINDy*.

Deep learning (DL) for SR systems works well on noisy data due to the general resistance of neural networks to outliers. An example of a Deep Symbolic Regression (DSR) system is proposed by[48], which is built for general SR tasks rather than specifically for data from the physical domain. This DL-based model uses reinforcement learning to train a generative RNN model of symbolic expressions. Furthermore, it adds a variation of the Monte Carlo policy gradient technique (termed "risk-seeking policy gradient") to fit the generative model to the precise formula.

SR systems based on genetic algorithms (GA) can efficiently enforce prior knowledge to reduce the search space of possible functions. For example, SR can adhere to a specific shape of the solution [49–52], or utilize probabilistic models to sample grammar of rules that determine how solutions are generated [53–56]. A simple yet effective implementation of GA for SR is *gplearn* Python Library[57]. It begins by building a population of naive random formulas representing a relationship between known independent variables (features) and targets (dependent variables) as tree-like structures. Then, in a stochastic optimization process, it performs replacement and recombination of the sub-trees, evaluating the fitness by executing the trees and assessing their output, and stochastic survival of the fittest. This method performs well on linear real-world problems[37] and can be easily manipulated as a base for more complex systems.

**Knowledge integration for SR systems.** Physical models must adhere to first principles and domain-specific theoretical considerations. From the perspective of the SR task, the search space should be reduced from all possible combinations into a space of solely the equations that comply with the physical restraints. Multiple knowledge integration methods have been proposed, which can be roughly divided into three main groups.

The first group of methods, the structure-related search space reduction, examines the structure of plausible equations, mainly by their partial derivatives, and incorporates assumptions (i.e., constraints) about them. For example,[58] suggested that physical models guarantee monotonic behavior concerning some of its features and narrow the search space to include only monotonic functions. Extending this line of thought,[59,60] suggested adding knowledge about convexity instead of only looking at monotonicity. Unlike monotonicity, the addition of convexity constraint does not stem from physical reasoning but rather from the necessity to formulate physical models as functions that can be optimized efficiently.[59] also suggested narrowing the search by assumptions about symmetry, as they suggested that all physical models are expected to be symmetric concerning the order of their arguments. When holds, these assumptions are shown to make the SR search much more efficient.

The second group of methods, the physical laws search space reduction, emphasizes the fundamental laws that any feasible solution should comply with. For example,[61] constrain the search space with conservation first principles. In contrast, the SR system *LGGA*[62] reduces the search space with more specific physical knowledge formulated as mathematical constraints. Another example is the *Multi-objective SR system* for dynamic models[63], which considers knowledge about steady-state characteristics or local behavior to direct the search efforts towards a logical result.

The third group of methods is a combination of the first two. Here fundamental knowledge is expressed as simple mathematical or logical constraints that can be bounded to specific parameters or ranges. Opposed to the structure-related methods, the user gains the power to choose if and where to apply any function-structure assumption. Nonetheless, as opposed to the physical law's methods, the user loses the ability to apply complex laws such as the first principles. For example, *SRFC*[64] uses GA to verify the candidate solutions by the given structural constraints such as symmetry, monotonicity or convexity, or knowledge constraints such as logical range of the result, its slope, or boundary conditions.

In light of the importance of integrating knowledge in SR tasks and the vast potential of knowledge loss during execution[65], we believe it is essential to construct the SR pipeline with the scientist at its core.

**State-of-the-art SR systems.** The current state-of-the-art SR system in physics is the so-called *AI Feynman* system [66]. *AI Feynman* combines neural network fitting with a recursive algorithm that decomposes the initial problem into simpler ones. Meaning that, suppose the problem is not directly solvable by polynomial fitting or brute-force search. In that case, *AI Feynman* trains a NN on the data to estimate the structure of the function by five simplifying properties presumably existing in it (i.e., Low-order polynomial, compositionality, smoothness, symmetry, and separability). If simplifying properties are detected, they are exploited to simplify and solve the problem recursively. Additionally, if dimensional samples are provided, a dimensional analysis solver is applied, doubling as a feature selection method that reduces the search space of the unknown equation. This is done by constructing a new set of non-dimensional features containing at least one representation of each dimensional (original) feature, and the smallest number of non-dimensional features possible. An updated version of the algorithm adds Pareto optimization with an information-theoretic complexity metric to improve robustness to noise[37,67].

This vast exploitation of simplifying properties enabled *AI Feynman* to excel at detecting 120 different physical equations, significantly outperforming the preexisting state-of-the-art SR for physical data. However, we argue that *AI Feynman* uses a series of restrictive assumptions that might lead to indefinite failure in cases outside the Feynman dataset. First, physical mechanisms might be implicit, therefore, undetectable if separability is assumed (e.g., the equation can presumably be written as a sum or product of two parts with no variables in common). Examples of such implicit functions in physics may be linkages behavior in mechanical engineering[68], or motion in fluids with a non-linear drag force[69]. Second, the application of automatic dimensional analysis does not allow the construction of specific non-dimensional numbers that are known to be related to the target or suspected of it. Therefore, it denies the integration of valuable domain knowledge that may reduce the search space or direct the search efforts in the right direction. Another shortcoming of this system is its high sensitivity even to small amounts of noise[37], making it hard to implement on real-world measurements.

Recently,[37] introduced *SRBench*, a benchmarking platform for SR that features 21 algorithms tested on 252 datasets, containing observational data collected from physical processes and data generated synthetically from static functions or simulations. The authors revealed that *Operon* by[70] was the best-performing framework in terms of accuracy. In contrast, *GP-GOMEA* by[71] was the best-performing framework in terms of the simplicity of the found mathematical expressions. Both frameworks, like most of the SR frameworks examined in *SRBench*, were not constructed to work specifically in the physical domain, meaning they do not constrain the outcome by any physical reasoning like dimensionality or monotonicity. In contrast, physics-oriented SR frameworks that were presented in *SRBench*, such as the strongly-typed GP[72], grammatical evolution[73], and grammar-guided[74] frameworks, were shown to perform, on average, worse than the other models[37]. This might be because most of the examined datasets were not physical, meaning that the physics-informed SR systems had a disadvantage.

Based on the conclusions from *SRBench*, we decided to compare SciMED to two SR systems: *AI Feynman*, which is the state-of-the-art for physical purposes, and the general SR framework of *GP-GOMEA* that excels in finding accurate yet simple mathematical models.

Like in traditional regression attempts, *GP-GOMEA* prioritizes human interoperability of the resulting equation. To do so, *GP-GOMEA* prevents bloat by implementing a strict constraint on equation length. In addition, to maintain the accuracy of the equation, the system estimates what patterns to propagate. This is done by performing variation based on the linkage model, to capture genotypic interdependencies. Because a short yet accurate equation is needed in many physical use cases, *GP-GOMEA* has been adapted in several physical regression efforts[75,76]. This is despite the fact that it does not consider any domain knowledge or physical requirements aside from interoperability.

### Scientist-Machine Equation Detector

In this section, we introduce Scientist-Machine Equation Detector (SciMED) for finding symbolic expressions from physical datasets with the SITL approach. SciMED is a free open source library in Python available through https://github.com/LironSimon/SciMED. SciMED is constructed from four components: GA-based a-priori feature selection, GA-based automatic machine learning (AutoML), GA-based symbolic regression, and Las Vegas (LV) search symbolic regression, as illustrated in Fig. 1. Each component allows the user to easily insert physical knowledge or assumptions, directing the search process toward a more credible result with fewer resources.

In particular, the a-priori feature selection component is applied at the user's discretion in cases where the user wants to select the most informative feature out of a group of features contributing the same information in essence (for example, select between measurements of the core, boundary or average temperature of a body). This selection process is not to be confused with the selection process performed by an SR component. Furthermore, we use an AutoML component to facilitate the SR task by enriching the data with synthetic samples. If the a-priori feature selection component is applied, the AutoML component also functions as its fitness function. Additionally, we use two approaches for SR. The GA-based one is less resource and time-consuming but stochastic, which may result in sub-optimal results. In contrast, the LV-based component is more computationally expensive but more stable and accurate, on average. A user can decide whether to use the GA, LV, or both SR components for a given task. As a default, the GA-SR is applied 20 times, and only if its results do not pass the criteria for accuracy or stability, the LV-Sr is initiated.

Below is a detailed description of each component and their interactions, and an overview of the SITL integration points in SciMED's pipeline.

**A priori feature selection component.** The number of parameters needed for an SR task proliferates, especially in non-linear problems with an unknown model. Like in traditional SR, the choice of which parameters to include can dramatically affect the result. Therefore, one must balance between retaining all relevant information and not obscuring the dynamics by creating too big search space. In this component, we offer a novel method for testing several hypotheses about informative parameters without increasing the search space for the SR.

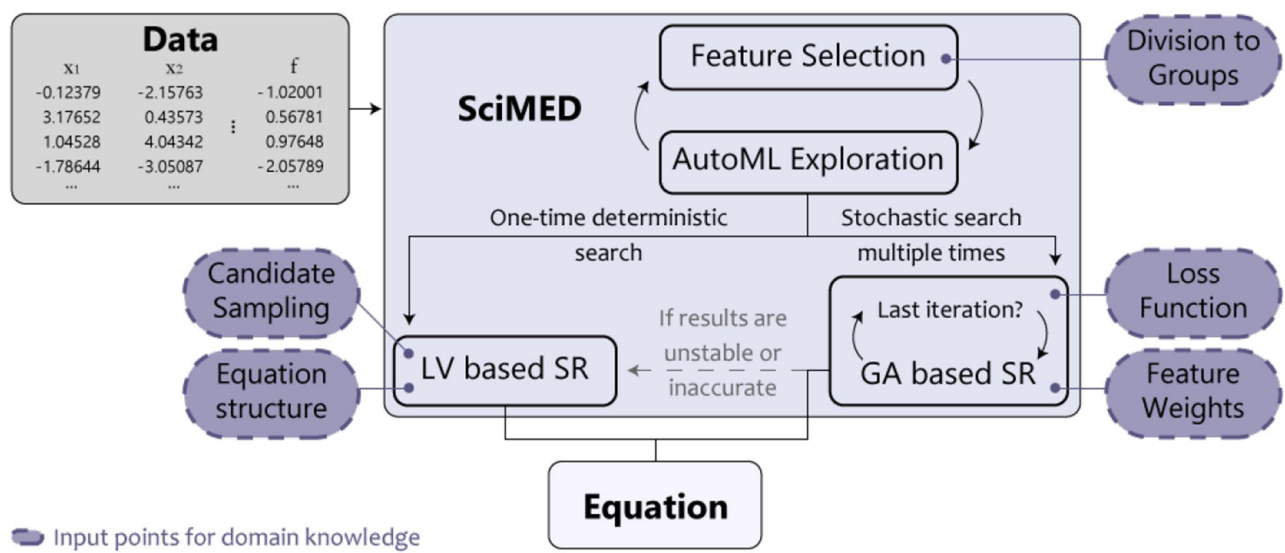

**Figure 1.** A schematic illustration of SciMED's structure. It is constructed from four components that each can be independently switched off or on. Theoretical knowledge or hypothesis can be entered at five input junctions, affecting the equation SciMED finds.

Here, users can suggest various plausible representations of dimensional or non-dimensional features based on knowledge or an educated hypothesis. But, since there is more than one way to acquire knowledge of a single physical attribute (i.e., feature), the user might want to explore various plausible representations of that attribute while keeping in mind that all representations contribute the same knowledge in essence. In such a case, the user can declare distinct groups of features, where each feature contributes the same knowledge. The allocation of features to groups is done by providing SciMED with the data as a table, and meta-data of the specific ranges of adjacent columns corresponding to each group. If no meta-data is introduced, SciMED assumes that each feature is the sole representation of a distinct group.

Formally, let $\mathbb{F} := \{f_1, \ldots, f_n\}$ be the set of provided features to SciMED and $\Phi := \{\phi_1, \ldots, \phi_l\}$ a set of features sets such that $\bigcup_{i=1}^{l} \phi_i = \mathbb{F}$ and $\forall i, j \in [1, \ldots, l] : \phi_i \cap \phi_j = \emptyset$ where $i \neq j$. Thus, given $\mathbb{F}$ and $\Phi$, SciMED chooses only a single feature for each set $\phi_i$ for $i \in [1, \ldots, l]$.

We implement this behavior using a GA-based approach as follows. A *chromosome* is defined by the subset of features from the feature pool ($\mathbb{F}$), where each *gene* in the chromosome is a feature from a distinct group ($\phi$). This information is encoded using a $n$-bit vector, where the $i_{th}$ bit in the vector corresponds to the $i_{th}$ feature group and represents which feature from this group was selected. This ensures each chromosome contains precisely one feature from each group the user provides, as needed.

Without *a priori* knowledge provided by the user, the GA is designed to optimize two objectives: a) maximize the obtained model's accuracy (or any other fitness metric used by the user), and b) minimize the number of features selected. To do so, we define the following fitness function:

$$FS_{\text{fitness}}(S) := \omega_{fs} M_{\text{fitness}}(S) + (1 - \omega_{fs}) |S|/|\mathbb{F}|, \tag{1}$$

where $\omega_{fs} \in [0, 1]$ is a balance weight between the model's performance and the obtained feature subset's size reduction ($|S|/|\mathbb{F}| \in [0, 1]$) and $M_{\text{fitness}}(S)$ is the model's fitness function outcome (between 0 and 1) for a chosen feature subset $S$. Of note, the fitness function is chosen by the user according to the task at hand. For instance, for regression tasks, a mean square error metric is a possible candidate.

In addition, the three genetic operators: selection, crossover, and mutation, are defined as follows. First, the *selection* operator implements a "tournament with royalty" operator[77], where each chromosome has a probability of being chosen for the next generation corresponding to its normalized fitness function. While the chromosomes in the top $\delta \in [0, 1)$ portion are chosen at least once for the next generation. The fitness score for each chromosome is assigned using the AutoML component, described in the next chapter. Second, the *crossover* operator is the "single-point" crossover operator[78], where a point $i$ is chosen at random so that the first $i$ bits are from one parent and the remaining bits are from the second parent. Lastly, the *mutation* operator mutates a chromosome in each individual with a probability $\rho$ determined by the normalized size of the feature group it represents. If a specific feature in the chromosome is chosen, it is randomly altered to represent a different feature from the same group.

Figure 2 demonstrates an example of the feature selection process, where the dataset is divided into nine feature groups, using the knowledge provided by SITL; four groups contain seven features, and five groups contain only one feature (hence they do not undergo a selection process). After the a-priori feature selection process is completed, a dataset of only nine features (equal to the number of groups) proceeds to the SR.

**Automatic machine learning extrapolation component.** In this component, we train an ML algorithm to perform "black-box" predictions of the target value. This is used to generate synthetic data from the sampled data, in order to cover the input space for the SR task uniformly. The motivation for that is that insufficient input space coverage is one of the leading challenges of applying SR methods on experimental data[79]. Additionally, suppose the *a priori* feature selection component is applied. In that case, this component assigns fitness scores to each chromosome corresponding to the accuracy achieved on the subset of features dictated by the chromosome.

Formally, given a dataset $D \in \mathbb{R}^{z,k}$ with $k \in \mathbb{N}$ features and $z \in \mathbb{N}$ samples, we utilize the TPOT[80] AutoML library, that uses a GA-based approach, to generate and test ML pipelines based on the popular scikit-learn library[81]. Formally, we run the TPOT regressor search method with a computational budget limitation to obtain an ML pipeline that will approximate and generalize the data. To prevent overfitting, the *k-fold* method is used[82]. Moreover, we allow the model's performance to be a vector of metrics (for instance [MAE, MSE, $R^2$, T-test's $p$ value]), computing the Pareto front's integral[83] to obtain the final model's performance score. Once the model is obtained, the mean and standard deviation of the $k$ iterations are computed and tested against threshold values provided by the user, $\zeta_{mean}$ and $\zeta_{std}$, respectively. If the model is extrapolating the data well and stable enough across the data, as reflected by these two values, additional $\tau \in \mathbb{N}$ synthetic values are sampled by querying the obtained model. The user defines the distribution and value of $\tau$. As a default, the user provides a radius $r \in \mathbb{R}^+$ and several neighbors points $\kappa \in \mathbb{N}$ such that each syntactic data point is of distance $d \leq r$ of at least $\kappa$ data points. Intuitively, these two parameters ensure that the synthetic data enrich the input domain while not introducing too far values for which our confidence is low.

In addition, this component of SciMED sheds light on the feature groups selected by the user[84–86], providing physical insight before obtaining symbolic expressions.

**Genetic algorithm based symbolic regression component.** In this component, we utilize the GA approach to find a symbolic expression. In particular, we follow the work proposed by[87], which extends the *gplearn* library[57]. Formally, a chromosome is represented using the S-expression[88], mixing between variables, constants, and functions. Initially, we use the *Full initialization* method, where all the S-expressions represented trees in the first generation have all their leaf nodes (variables or constants) at the maximal distance from the root. Afterward, and for each generation, the three GA operators are implemented as follows:

- Selection–a combination of the *tournament with royalty* and the *Genitor* methods. Namely, the chromosome in the population are ranked by their fitness score. A portion of the population with the best fitness scores is carried forward into the next generation. Afterward, from the remaining chromosome, a chromosome is likely carried forward into the next generation by a probability relative to its fitness score, normalized to the sum of all the fitness scores in the population. When a chromosome is chosen to be included in the next generation, several mutations are performed corresponding to its normalized fitness score.
- Mutation–we use the *point mutation* method. Namely, a node in the S-expression tree is chosen randomly and replaced with another feasible value. In the context of SR, parameters are replaced with parameters, and functions are replaced with functions of the same number of arguments.
- Crossover–firstly, two chromosomes are randomly taken from the population. A random subtree of the first chromosome is then replaced with a subtree from the second chromosome program, and the other way around, to generate two new chromosomes.

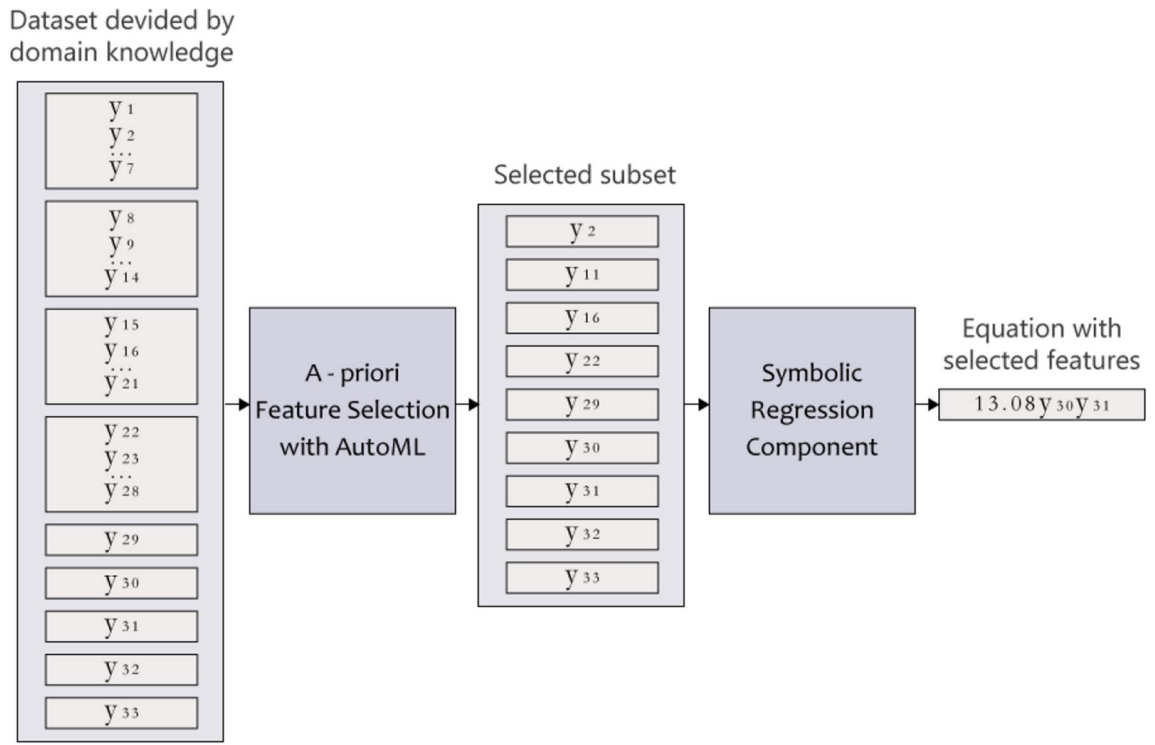

**Figure 2.** An illustration of the feature selection process. One can see how SciMED performs feature selection for experiment B, where all features and the outcome represent the actual process and desired result. Here, a dataset with 33 features ($f(\vec{y}) = (y_1, y_2, \ldots, y_{33})$) is divided to 9 feature groups, using the meta-data provided by the user, where a single feature is selected from each group. This subset is passed to an SR component, revealing the unknown equation containing only two features. The physical background of the features and the division into groups is explained in detail in the Appendix.

After applying these three operators, the evaluating phase takes place to determine the fitness score of the new chromosome population for the next generation. At this step, the chromosome is evaluated in a *k-fold* manner[82] and takes the value of the evaluation of the whole data rather than a specific section. Each evaluation of the chromosome on the dataset is done with a loss function provided by the user. As default, we use the following loss function, inspired by[89], that gets a chromosome ($g$), the gene's predicted value ($y_p$), and the target's true value ($y_t$):

$$L(g, y_p, y_t) := \omega_1 \|y_p - y_t\|_1 + \omega_2 \|y_p - y_t\|_2 + \omega_3 \|y_p - y_t\|_\infty + \Psi|g|, \quad (2)$$

where $\{\omega_i\}_{i=1}^3 \in [0, 1]$ such that $\sum_{i=1}^3 \omega_i = 1$, $\|z\|_x$ is the $L^x$ norm of the vector $z$, $\Psi \in \mathbb{R}^+$ is a weight for punishing the chromosome proportional to its size. It is important to note that SciMED allows fine-tuning the value of $\Psi$ over a range using the grid-search method[90].

This component can be initialized several times with different initial populations. After all the runs are finished, the stability of the outcomes is tested in two ways. First, the standard deviation of a selected performance metric is evaluated to identify whether it's converging. Second, equations from each run are compared to check whether a particular equation is repeated in a minimum $\chi \in [0, 1]$) percentage of the runs ($\chi$ vs user-defined).

**Las Vegas symbolic regression component.** In this component, we search for a symbolic expression in a more stable but computationally expensive way than in the GA-based SR component. Similar to the GA-based SR component, we define a candidate solution to be represented by an S-expression tree. However, in this component, candidate functions that get more than two inputs are divided into an S-expression tree of functions that do satisfy this condition. Similarly, all single-input candidate functions are rephrased to get two inputs such that the second input is never used. This allows us to represent all candidate solutions as full binary trees (FBT). As such, given a range of candidate solution sizes $\xi_1, \xi_2$ ($\xi_1 \leq \xi_2$), we compute all possible FBTs. Then we randomly allocate functions and variables for the FBT, and evaluate it using Eq. (2). During the evaluation process, the outcome of the FBT computation is used to train a linear regression to find coefficients of the obtained symbolic expression. After a pre-defined number of such candidate solutions are evaluated ($\theta \in \mathbb{N}$), we update the probability that a new sample would be chosen by setting it to be the normalized value of K-nearest neighbors[91] from the already sampled candidate solutions, inspired by[92]. This process is terminated once a user-defined number of attempts (or given computation time) has been reached, $\rho \in \mathbb{N}$, or if all possible allocations were evaluated. Either way, the candidate solution with the smallest loss value is returned. The motivation for this approach is to find and sample the most promising areas in the search space iteratively.

A user may opt to direct the stochastic search process by introducing two types of knowledge: a) a hypothesis for the structure of the optimal solution, and b) a sampling strategy believed to obtain an optimal solution faster.

Notably, as this component is computationally heavy, a user can avoid it entirely and settle for the previous GA-based SR. Furthermore, this waiver can be automatically implemented if the results obtained from the GA-based SR are consistent, appearing over a pre-defined portion of the outcomes from multiple runs and maintaining the T-test's $p$ value and coefficient of determination ($R^2$) of over a user-defined percentage, $\chi$.

**Scientist-in-the-loop integration points.** SciMED's design allows the user to integrate knowledge into the search and optimization process in five distinct but related places, as shown in Fig. 1, alongside a reach set of hyperparameters that can be customized to direct the search efforts and their computational resources.

The five knowledge integration points in SciMED's design are:

1. Division into distinct feature groups: Providing the feature groups, $\Phi$, allows SciMED to search for the most informative subset of features such that each feature belongs to one group. This allows the user to examine multiple hypotheses or educated guesses without increasing the search space for the SR, eliminating the need to solve Eq. (1).
2. Loss function selection: for the GA-SR component. Provides the user with the option of replacing our loss function (Eq. (2)) with one that is believed to provide a better fit for the specific SR task.
3. Allocation of feature weights: for the GA-SR component. Suppose the user believes one feature is more likely to play a role in the dynamics than another. In that case, he can allocate more weight to it to increase its chance of being included in the next generation of candidate functions. Then, the GA algorithms would stochastically spend more resources to try candidates with this feature. A user can also assign less weight to a feature to reduce its likelihood of being included in candidate functions.
4. Candidate sampling strategy: for the LV-SR component. Users can direct the random sampling of the FBT's topology and allocations. Similar to the allocation of feature weights in the GA component, the user may direct the search effort by enforcing that the LV-SR component will sample FBTs with specific features for $x$% of the time, and the remaining options for the rest $(100 - x)$%. In addition, the user may allocate more resources toward examining FBTs of a specific size. Hence, for example, the user suggests that the solution can be represented with a FBT of size between 3 and 13. However, the user believes it is probably either 7 or 9. Thus, the user can enforce that FBTs of these sizes would be sampled $x$% of the time, and FBTs of other sizes for the remaining time.
5. Equation structure constraint: for the the LV-SR component. Users can enforce constraints on the FBT's topology or feature allocation. For example, if the user is confident that the final solution should contain a specific feature, he can use an allocation constraint to narrow the search space. Another option is to condition all candidate functions by topology and allocation assumptions, enforcing a sub-FBT that must be included in them. For instance, search only FBTs that have feature $f_1$ and feature $f_2$ with a sum function between them (e.g., $f_1 + f_2$).

It is encouraged to provide SciMED with non-dimensional datasets, meaning that the user performed the dimensional analysis independently. This is because it guarantees that the units of the target variable agree with the units obtained by the solution. Additionally, during the independent analysis, the user can construct non-dimensional ratios known or suspected to be informative about the target that might not result from an automatic analysis.

Furthermore, SciMED SciMED has many hyperparameters that can affect its performance. Table 1 lists and describes these parameters and their default values. Below, we provide guidance on choosing appropriate values for several key hyperparameters.

First, the number of folds in the cross-validation ($k$) is a critical parameter in the AutoML component. Increasing $k$ improves the performance of each model, but reduces the number of ML pipelines that can be evaluated in the same amount of time. Therefore, the value of $k$ should be chosen based on the size of the dataset. For example, values of 5 or 10 are often used because they provide a good balance between computational time and evaluation accuracy in many cases[93].

Second, $\tau$, the number of samples added by the data extrapolation performed by the AutoML component. A value of $\tau$ that is too small does not contribute much to the other components while just slightly increasing the computation time. However, a large value of $\tau$ may result in drift, and the connections detected by the AutoML component would override the original connections inside the data. As such, $\tau$ should be a relatively small portion of the original dataset. Recent work shows that integrating up to 25% of synthetic data obtained by an ML model or generative adversarial neural networks can contribute to classification and regression tasks[94–96]. Following these results, we suggest setting $\tau$ between 5% and 20% of the original data set size.

| Hyperparameter | Description | Default value |
|---|---|---|
| Dimensional or non-dimensional data | A Boolean parameter that indicates to SciMED if the data is physically dimensional or not | False |
| $(\Phi)$ | A set of feature sets that is used by the GA feature selection component to choose a single feature from each set | Per experiment |
| $\omega_{fs} \in [0, 1]$ | A balance weight between the model's performance and the chosen feature set's size | 0.9 |
| Feature selection fitness function ($M_{fitness}$) | A function that accepts a feature set, a model's prediction vector, and a ground true vector and returns the fitness (performance) of the model between 0 and 1 | $L_1$-normalized mean absolute error |
| $\delta \in [0, 1)$ | The portion of the $\delta$ most fitted chromosomes that are taken to the next generation in the feature selection's component | 0.05 |
| $k \in \mathbb{N}$ | The automatic machine learning component's k-fold cross validation | 5 |
| $\zeta_{mean}$ & $\zeta_{std} \in \mathbb{R}$ | The mean and standard divination automatic machine learning component's threshold to use the data extrapolation component | 0.1 & 0.02 |
| $\tau \in \mathbb{N}$ | The number of samples added by the data extrapolation component | 10% of the original sample's number |
| $r \in \mathbb{R}$ | The maximum Euclidean distance from $\kappa$ neighbors a synthetic sample required to be in, used by the data extrapolation component | 0.025 |
| $\kappa \in \mathbb{N}$ | The number of neighbors required to be in a distance $r$ from a synthetic sample, used by the data extrapolation component | 3 |
| $\{\omega_i\}_{i=1}^3 \in [0, 1]^3$ | is the weights of the three terms in Eq. (2) | [1/3, 1/3, 1/3] |
| $\Psi \in \mathbb{R}$ | A weight for punishing the chromosome proportional to its size in the GA-SR component | 0.05 |
| $\chi \in [0, 1]$ | The minimal portion of repeated runs of the GA-SR component that resulted in the same equation to declare a success | 0.5 |
| $\xi_1$ & $\xi_2 \in \mathbb{N}$ | A range of candidate solution sizes for the LV-SR component | 3 & 13 |
| $\theta \in \mathbb{R}$ | Number of FBTs candidate solutions evaluated before updating the sampling distribution | $10^3$ |
| $\rho \in \mathbb{N}$ | The maximum number of FBTs candidate solutions evaluations | $10^7$ |

**Table 1.** A description of SciMED's hyperparameters and their default values.

Third, the range $(\xi_1, \xi_2)$ in the LV-SR component, that is responsible for the FBT's topologies size. One can notice that if the topology size of the optimal solution $c$ is not between $(\xi_1, \xi_2)$, it would not be obtained. Thus, $(\xi_1, \xi_2)$ should be large enough to capture such an optimal solution while not too large to avoid enormous search space that might result in expensive computation. A rule of thumb that one can follow is to look at the FBT's representation of other equations that stand at the base of the same physical domain[97,98].

The remaining hyperparameters values, shown in Table 1, are obtained using a trial-and-error approach, and tested on various equations and datasets, not including the ones included in this paper.

## Results

**Experimental design.** We evaluated SciMED on seven different experiments, testing its competitiveness against *AI Feynman* and *GP-GOMEA* on highly noisy data, demonstrating the contribution of knowledge integration, and evaluating its resistance to noise.

The first five experiments are designed to highlight the importance of different components in SciMED. First, we assessed SciMED ability to detect linear relations between features from scarce and noisy data (experiment A). Here, we aim to highlight the contribution of the GA-based SR component and its ability to perform SR efficiently. Second, we tested the ability of SciMED to find a linear equation from a vast dataset of tens of features (experiment B). This experiment aims to demonstrate the contribution of the *a priori* feature selection component by incorporating domain knowledge and reducing the search space. Third, we examined the ability of SciMED to find a non-linear equation from data with noise and a large number of features (compared to the average number of features in a benchmark set of 100 physical equations[66]). This experiment (experiment C) is intended to demonstrate the contribution of the LV-based SR and its robustness to noise. Here we highlight inferring the correct numerical value of the prefactor. This result means SciMED could correctly estimate the gravitational acceleration from noisy data, a difficult task by itself[99]. In experiment D, we demonstrate how the AutoML component may alert the user that a parameter of crucial importance is missing. To do so, we evaluated SciMED on a dataset with non-linear feature relations that is missing one essential feature. This experiment mimics a reasonable scenario in scientific research, where a researcher assumes to know all the parameters governing a phenomenon but neglects to consider (at least) one. To increase the difficulty of this experiment, the chosen feature has hidden physical relations to other introduced features. In turn, this may lead to misleading performance scores and highlights the difficulty of obtaining a reliable symbolic expression. In experiment E, we compare SciMED's to two state-of-the-art systems using a dataset of noisy measurements with 12 features. Here, only four features need to be selected to formulate a correct equation, but the high noise levels make SR difficult as all the features multiply with one another, increasing the noise in the target value significantly.

Later, in experiment F, we show the benefits of integrating domain knowledge into SciMED. We apply SciMED on experiment E with all binary combinations of domain knowledge integration (with or without knowledge insertion at each of the five input junctions). This results in $2^5 (= 32)$ possible configurations to run SciMED. As previously described, not all the information we provided directs the search toward the results, as might occur in real cases. The reasoning for the bits of knowledge inserted is listed in that chapter.

As a final experiment (experiment G), we performed robustness or noise analysis, demonstrating SciMED's performance in the presence of three different types of noise and at various noise levels.

**Experiments A-E: Comparison of SciMED to *AI Feynman* and *GP-GOMEA*.** For every experiment, we present three results: 1) A scatter plot of AutoML predictions versus ground truth. 2) A vector of performance scores for all components of SciMED. 3) The discovered equation. The discovered equations from each experiment were compared to those that were found by *AI Feynman* and *GP-GOMEA* systems.

Fig. 3 shows the prediction capabilities from the AutoML component in each experiment. For each plot, a linear regression line is fitted to the values predicted by the ML (noted as $f_{pred}$) as a function of the true target values (indicated as $f_{true}$). All coefficients of determination ($R^2$) scores indicate the found ML was accurate enough to enrich the data domain reliably. Specifically, in experiment B, the linear regression was optimal (i.e. $f_{pred} = f_{true}$). This makes sense, since experiment B did not include noise. In the other experiments with noise, a small number of outliers were observed. Specifically in experiment E, where the multiplication of noisy features significantly increases the percentage of noise in the target, predictions of higher targets had a larger error.

In experiment D, where a single variable was removed from the data, the accuracy of the predictions declined, as seen by both the $R^2$ score and the number of outliers. However, the decline was not as severe as expected because the removed variable depended on other variables given to SciMED, meaning it might have revealed the necessary information from the provided data.

Table 2 reports the performance scores achieved in experiments A-E by the AutoML and both SR components. Due to the multiple iterations of the AutoML and GA-based SR components, their scores are reported as a mean ± standard deviation. Following Fig. 3, one can see that the AutoML component shows good performance over all four metrics (MAE, MSE, $R^2$, T-test's $p$ value). As expected, the LV-based SR component consistently outperformed the GA-based SR component, presenting excellent results for all but experiment D.

For experiment D, the combination of good performance overall metrics by the AutoML component and poor performance overall metrics by the LV-SR component indicates that at least one dependent variable needs to be added to the dataset. This is because, on the one hand, the performance scores suggest that AutoML accurately learned the necessary information from the given variables. However, on the other hand, the robust SR component failed to find an equation that remotely describes the data (as seen by the zero-valued T-test's $p$ value). Hence no accurate equation can be formulated with the given variables, meaning at least one variable is missing from the equation.

In addition, Table 2 shows the advantages of combining the GA-based SR with the LV search; GA performs well on relatively simple SR tasks but fails when there is a vast search space or high levels of noise. Following that, it is evident that experiments B-E have higher MAE and MSE scores, coupled with lower $R^2$+ and T-test's $p$, compared to experiment A, which is more straightforward.

The SR results of SciMED are presented in Table 3 alongside those of *AI Feynman* and *GP-GOMEA*. In experiment A, all systems found the unknown equation despite the noise applied to the target. In experiment B, all systems correctly identified the two out of the 33 variables appearing in the equation and their algebraic relation. However, SciMED found a numerical prefactor smaller by 0.02 than the actual value and added a constant term of 0.03, compared to *AI Feynman* and *GP-GOMEA*, which found a prefactor with an error of 0.33 and 0.05, and did not add a constant term.

In experiment C, *AI Feynman* failed to find the correct equation, leaving out one parameter and incorrectly identifying the algebraic relationships and the numerical prefactor (identifying a prefactor smaller by 4.75 than the true value). SciMED and *GP-GOMEA* correctly identified the equation and its numerical prefactor with an error of 0.1 and 0.06. SciMED also added a small constant term of 0.04. In this experiment, the prefactor is linked to a physical constant - the gravitational acceleration *g* (for an explanation, see Appendix). Therefore, SciMED's identification of a prefactor within a 0.76% error means it could accurately learn the value of *g* used to construct the target from noisy data. This is considered a difficult task[99], that SciMED succeeded in.

In experiment D, where it is impossible to construct an equation for the target from the parameters in the data, SciMED resulted in an equation with the minimal MAE score it found, while *AI Feynman* and *GP-GOMEA* failed to terminate even after 12 computation hours with the Intel Core i7-1185G7 processor and Ubuntu 18.04 operation system. Those systems continuously added terms to the equation they tried to match the data. This indicates an advantage of SciMED, as the outcome with poor SR performance and good AutoML performance alerts the user to re-examine the data. On the other hand, *AI Feynman* and *GP-GOMEA* exhibited a common bloat issue that potentially leads to good performance scores by adding more terms to the equation but fails to generalize[100]. Of note, while alerting the user of potentially missing information is not unique to SciMED (other GA-based SR models have the same capability), it is an added value that SR models based on brute-force and sparse matrices do not have.

In experiment E, SciMED was the only one to find the correct features and algebraic relationships without domain knowledge. This shows its competitiveness with its default configuration.

**Experiment F: domain knowledge integration.** Here experiment E was repeated 32 times, each time withholding or adding information through a specific input junction, demonstrating the impact of domain knowledge.

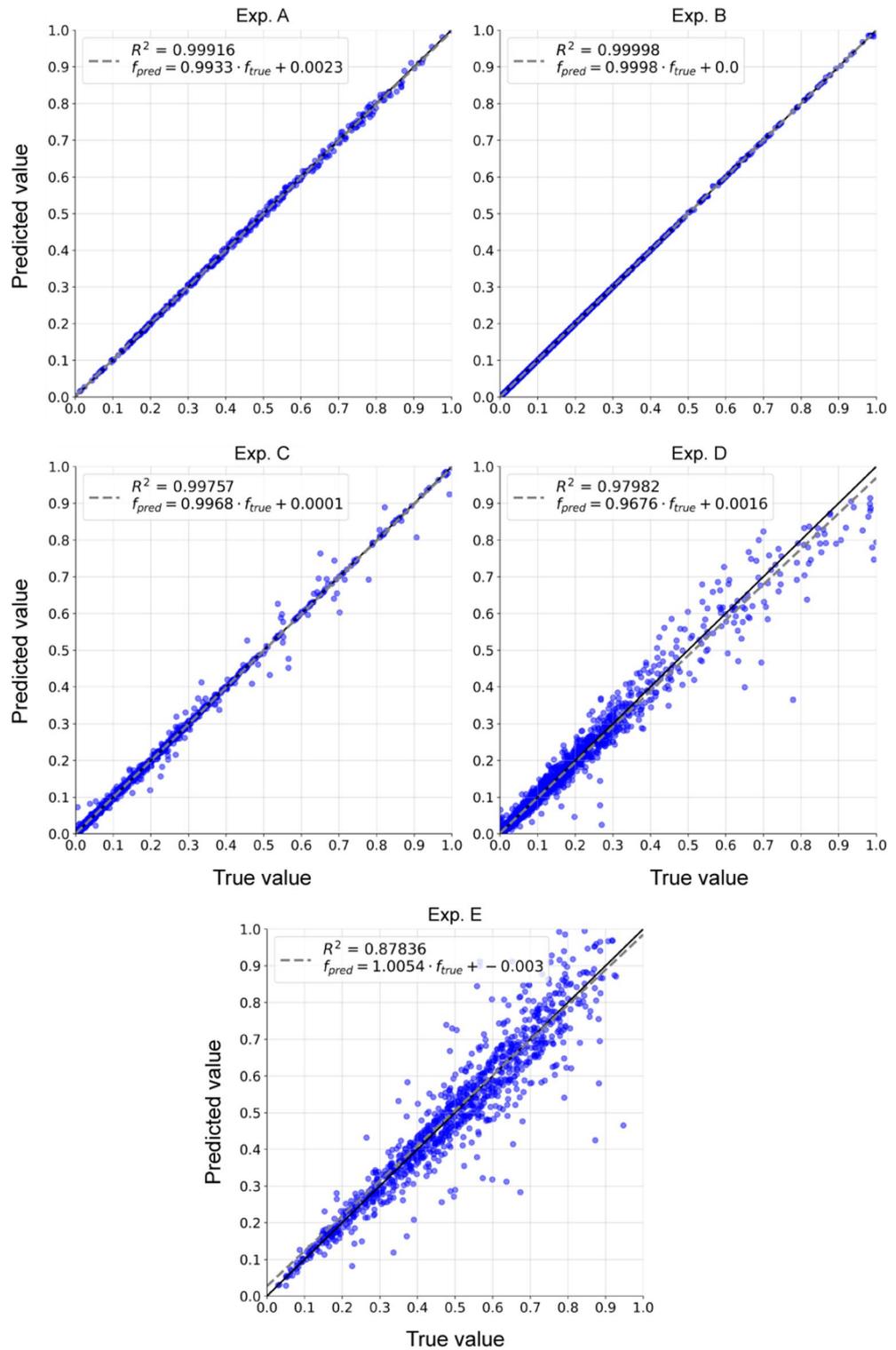

**Figure 3.** Predictions acquired with the ML pipeline found in the AutoML component versus the true target value. Lines represent the regression, and the respective equation is shown in the legend.

The knowledge inserted in this experiment is as follows: First, through the division to feature groups insertion point, the 12 features in the dataset were reduced to 10, using the *a-priori* feature selection component. Second, as Eq. (2) is considered well for a general case, an alternative loss function was used: $0.5\|y_p - y_t\|_1 + 0.5\|y_p - y_t\|_2 / \|y_p - y_t\|_\infty$. The motivation behind this loss function is that large errors have relatively small impacts, allowing the model to capture the main behavior of the dynamics and not be influenced by anomalies resulting from the noise. Third, four features received bigger weights in the GA-SR component.

| Experiment | Component | MAE | MSE | $R^2$ | T-test's $p$ value |
|---|---|---|---|---|---|
| A | AutoML | $0.006 \pm 0.001$ | $0.000 \pm 0.000$ | $1.000 \pm 0.000$ | $0.960 \pm 0.007$ |
|  | GA - SR | $0.457 \pm 0.000$ | $0.386 \pm 0.000$ | $0.999 \pm 0.000$ | $0.993 \pm 0.000$ |
|  | LV - SR | 0.439 | 0.371 | 0.999 | 0.980 |
| B | AutoML | $0.000 \pm 0.000$ | $0.000 \pm 0.000$ | $1.000 \pm 0.000$ | $0.992 \pm 0.007$ |
|  | GA - SR | $1.111 \pm 0.000$ | $13.979 \pm 0.000$ | $0.000 \pm 0.000$ | $0.000 \pm 0.000$ |
|  | LV - SR | 0.000 | 0.000 | 1.00 | 1.000 |
| C | AutoML | $0.005 \pm 0.001$ | $0.000 \pm 0.000$ | $0.986 \pm 0.004$ | $0.918 \pm 0.170$ |
|  | GA - SR | $1.036 \pm 0.000$ | $6.366 \pm 0.000$ | $0.000 \pm 0.000$ | $0.000 \pm 0.000$ |
|  | LV - SR | 0.002 | 0.000 | 0.989 | 0.994 |
| D | AutoML | $0.009 \pm 0.001$ | $0.000 \pm 0.000$ | $0.953 \pm 0.008$ | $0.948 \pm 0.047$ |
|  | GA - SR | $1.407 \pm 0.000$ | $8.134 \pm 0.000$ | $0.000 \pm 0.000$ | $0.000 \pm 0.000$ |
|  | LV - SR | 416.865 | 231,982.744 | 0.000 | 0.000 |
| E | AutoML | $0.011 \pm 0.003$ | $0.002 \pm 0.000$ | $0.937 \pm 0.021$ | $0.823 \pm 0.091$ |
|  | GA - SR | $0.977 \pm 0.008$ | $25.722 \pm 1.705$ | $0.838 \pm 0.046$ | $0.360 \pm 0.093$ |
|  | LV - SR | 0.014 | 0.006 | 0.878 | 0.758 |

**Table 2.** Performance scores of the AutoML and both SR components of SciMED, for experiments A-D. As the AutoML and GA-based SR components are run multiple times, their scores are presented as a mean ± standard deviation.

|  | Exp. A | Exp. B | Exp. C | Exp. D | Exp. E |
|---|---|---|---|---|---|
|  | $f_1(x_1, x_2, x_3)$ | $f_2(y_1, y_2, \ldots, y_{33})$ | $f_3(z_1, z_2, z_3, z_4)$ | $f_4 = f_3(z_1, z_2, z_3)$ | $f_5(h_1, h_2, \ldots, h_{12})$ |
| Target Eq. | $f_1 := x_1 + x_2 \cdot x_3$ | $f_2 := 1.33 y_{30} \cdot y_{31}$ | $f_3 := 13.08 \frac{(z_1-z_2)z_3}{z_2 \cdot z_4^2}$ | $f_4 := NA$ | $f_5 := 0.125 h_1 \cdot h_2 \cdot h_3^2 \cdot h_4^2$ |
| SciMED | $x_1 + x_2 \cdot x_3$ | $1.31 y_{30} \cdot y_{31} + 0.03$ | $12.98 \frac{(z_1-z_2)z_3}{z_2 \cdot z_4^2} + 0.04$ | $\frac{z_2}{z_1 \cdot z_3}$ | $0.121 h_1 \cdot h_2 \cdot h_3^2 \cdot h_4^2 + 0.07$ |
| AI Feynman | $x_1 + x_2 \cdot x_3$ | $y_{30} \cdot y_{31}$ | $8.33 \frac{(z_2+z_3)z_3}{z_2 \cdot 2z_4}$ | FT (50+ terms) | $0.233(h_1 \cdot h_3^2 \cdot h_4^2 + \frac{h_2 \cdot h_5^2 \cdot h_4^2}{h_1 + h_8})$ |
| GP GOMEA | $x_1 + x_2 \cdot x_3$ | $1.38 y_{30} \cdot y_{31}$ | $13.14 \frac{(z_1-z_2)z_3}{z_2 \cdot z_4^2}$ | FT (30+ terms) | $0.059(h_1 + h_6) h_4^2 \cdot h_3 (h3 - h_7)$ |

**Table 3.** The target equations for the experiments SciMED, *AI Feynmn* and *GP GOMEA* were tested on. The data for exp. D was generated using Eq. C, but a partial dataset was given to SciMED. Therefore there is no true function underlying the samples. Each system was forced to terminate (FT) its computation after 12 hours. For a physical representations of $\vec{x}, \vec{y}, \vec{z}, \vec{h}$ see the Appendix.

Two of them were, in fact, in the unknown equation ($h_3, h_4$), and the other two were not ($h_7, h_8$). This helped us demonstrate a more reliable scenario of knowledge integration, where sometimes the user's hypotheses aren't correct. Forth, the LV-SR search was set to focus on FBTs that contain a specific feature ($h_3$) for 80% of the search time, using the candidate sampling strategy input junction. Fifth, a constraint on equation structure was implemented to enforce that all FBT candidates include $h_1$.

Fig. 4 shows the percent of correct equations from 20 iterations of the same configuration, along with the normalized computational time for each case. The normalization was performed by dividing the time it took to complete each experiment by the time it took to complete the experiment with no added domain knowledge. As expected, the more knowledge added to guide the search, the more the success rate increases and the computational time decreases.

**Experiment G: noise analysis.** For the noise analysis, we repeated experiments A and C while increasing the percent of noise introduced to either the input variables, target variable, or both variables. We repeated experiments A and C for $n = 100$ times for each type and amount of noise, reporting the percent of correct equations from all results.

The results obtained from both SR components are presented in Fig. 5. As expected, the more complex the unknown equation is, the more sensitive to noise SciMED becomes, as revealed by comparing the two columns. In addition, the Las Vegas-based SR performed better on higher noise levels than the GA-based SR component for both cases, as revealed by comparing the results in the first and second rows.

### Discussion

This work presents SciMED, a novel SR system that combines the latest computational frameworks with a scientist-in-the-loop approach. This way, SciMED emphasizes knowledge specific to its current task to direct its SR efforts. It is constructed of four components that can be switched on or off independently to suit user needs. In addition, it allows users to easily introduce domain knowledge to improve accuracy and reduce computational time. To the best of our knowledge, allowing users to set distinct pairwise sets for the feature selection process is an unprecedented method of physical hypothesis evaluation that enables users to efficiently examine multiple hypotheses without increasing the SR search space. Thus, feature groups are a new and efficient way for researchers to explore several theories of the variables governing unknown dynamics that are otherwise unfeasible due to complex interactions between different feature groups.

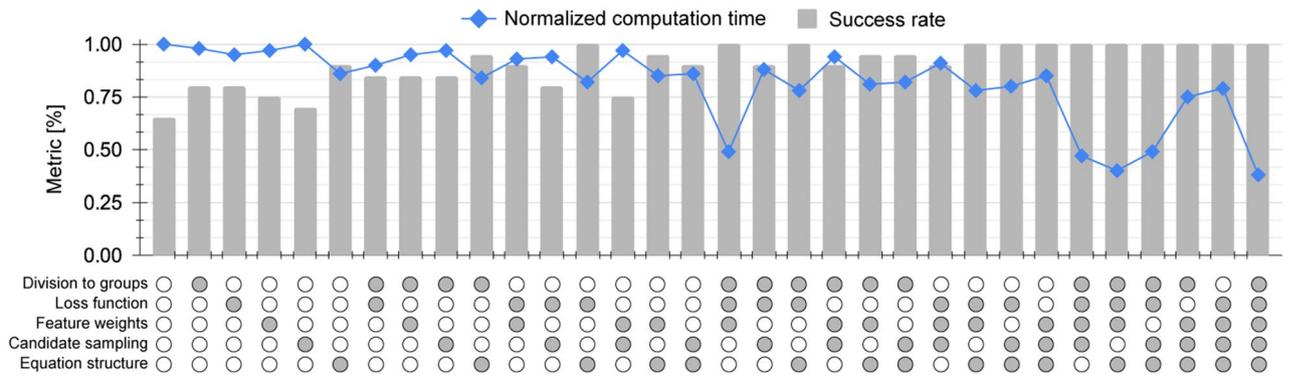

**Figure 4.** An analysis of the contribution of domain knowledge to the performance of SciMED. For each configuration of knowledge insertion, we conducted 20 iterations and recorded the percentage of correct results (grey bars) and the normalized computational time (blue scatter). The X-axis in the graph shows whether knowledge was inserted (grey) or withheld (white) from a specific input junction.

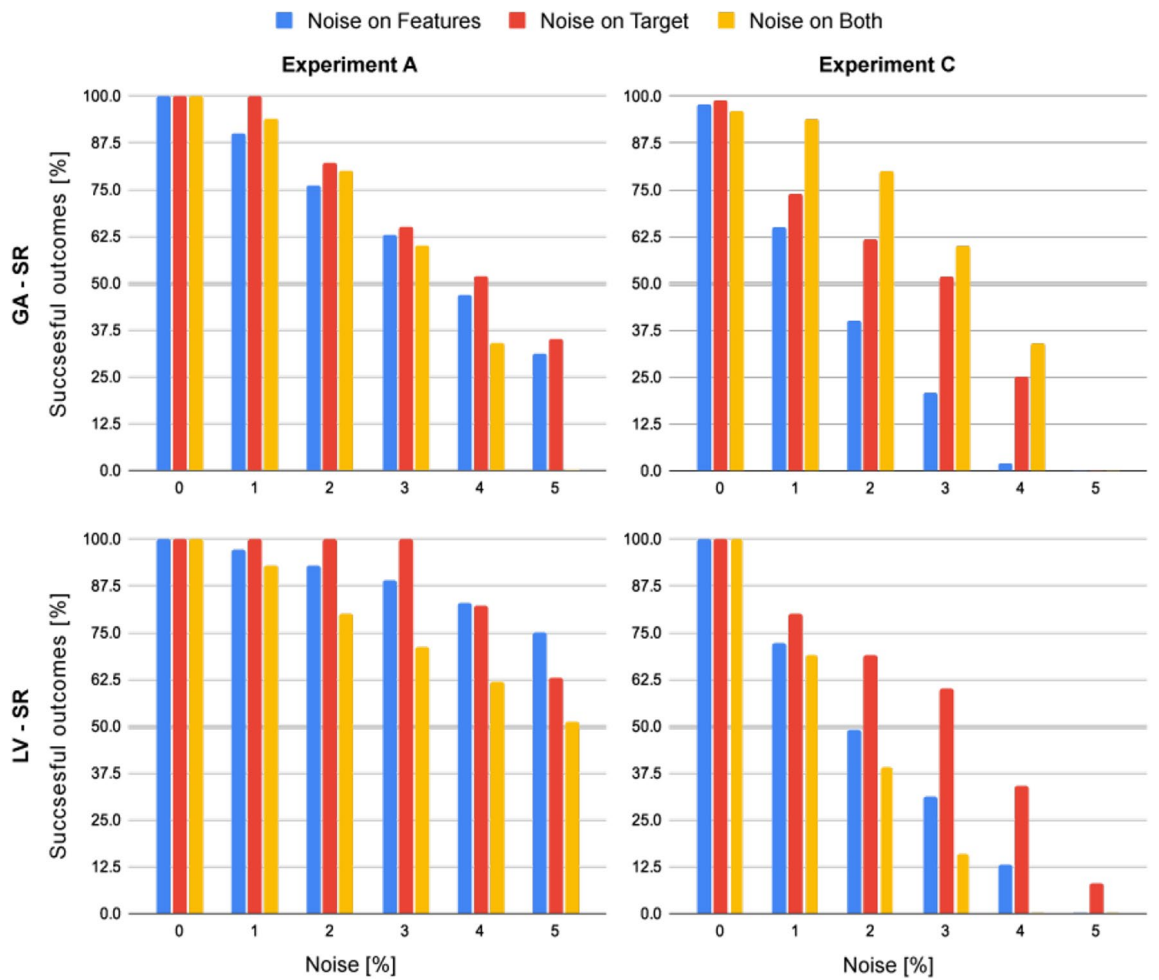

**Figure 5.** Noise analysis of SciMED displaying the percent of correct outcomes (equations) for each SR component. Results are obtained from $n = 100$ iterations for each percentage of noise added to the data of experiments A and C. The noise is divided into noise added to input variables (blue), noise added to the target variable (red), and noise to both (yellow).

To facilitate quantitative benchmarking of our and other symbolic regression algorithms, we tested SciMED, *AI Feynman* and *GP-GOMEA* on five cases, simulating real-world measurements with significant noise. We compare SciMED to them as *AI Feynman* is considered the cutting-edge system for physical purposes, while the general SR system of *GP-GOMEA* excels at finding accurate and straightforward mathematical models (a requirement for SR in the physical domain).

In the first two cases (experiments A-B), we highlight the contribution of the GA-based SR and feature selection components. For these cases, it is not surprising that *AI Feynman* also demonstrated good performance, as it brute-forces all the polynomials up to a fourth-order, including the two linear configurations of these cases. Nevertheless, in experiment B, SciMED slightly outperformed *AI Feynman* and *GP-GOMEA*, finding a more accurate numerical prefactor of the equation.

In the next two cases (experiments C-D), we emphasized the contribution of the LV-based SR and AutoML components. In experiment C, SciMED significantly outperformed *AI Feynman* by finding the correct equation within a 0.76% error of the numerical prefactor, compared to *AI Feynman* that converged to a false equation (as summarized in Table 3). Furthermore, the deduction of an accurate prefactor, linked to the gravitational acceleration constant, from data with Gaussian noise poses a known challenge to SR[99] that SciMED and *GP-GOMEA* succeeded at. Experiment D mimicked a scenario in which the user might fail to enter all the needed variables to explain the target. In such a case, an SR system should report its failure to converge to an equation of reasonable length rather than report a bloated equation of tens of variables that fails to generalize[100]. In this experiment, SciMED alerted the user that there is a possible dependent variable missing from the data and presented the equation with the lowest MAE score it found. *AI Feynman* and *GP-GOMEA* on the other hand, failed to terminate despite significant computational efforts. Instead, the systems reported an unreasonable equation of over 30 terms. Both results are shown in Table 3.

Finally, in experiment E, we highlight the competitiveness of the LV-based SR component by identifying a non-linear equation from a highly noisy dataset, that requires choosing 4 out of 12 features that range in similar values. In this task SciMED is the only system that accurately detects the correct features and their algebraic relation.

In experiment F, we perform SR on experiment E, this time adding different types of domain knowledge. The information provided helps SciMED to increase its success rate from 65% percent of the time to 100%, while decreasing the computational time expense. Fig. 4 demonstrates that each type of domain knowledge affects the success rate to a different extent, but all kinds of information (even if they contain partially incorrect assumptions) improve it.

We obtained the results of experiments A-F from data with noise introduced to the target variable to accede with previous work[16,66]. In practice, there are three types of noise one can experience in real-world data; noise in the target variable, noise in the input variables, and noise in both variables. The latter two, which were rarely presented in prior work (although common in practice), pose a more difficult challenge for SR as the amount of noise added to the target accumulates. A noise analysis on SciMED (experiment G) confirmed that: 1) SciMED is robust and withstands high levels of noise (compared to the levels tested in[16]) of all three types. 2) SciMED becomes increasingly sensitive to noise the more complex the unknown equation is (i.e., in terms of length or algebraic combinations of variables). 3) The Las Vegas-based SR performs better on higher noise levels than the GA-based SR component, meaning that the LV-SR component should be applied in case data is gathered with significant uncertainty.

One can wonder how applicative the SITL approach is in real-world scenarios, where the final result is unknown and a real discovery process is conducted. While integrating the correct guesses or knowledge may be considered more an art than science, knowledge integration has gained popularity in recent years[101] and has been well utilized by researchers and engineers[102,103]. Hence, it is expected that researchers could regularly use the SITL approach in their domain of expertise, or collaborate with others that can do so[104].

Although the presented results are promising, SciMED has limitations. First, since SciMED's main advantage is in the domain knowledge provided by the user, it is also its main limitation. Introducing false hypotheses may reduce the search space too much, making it more complicated or impossible to deduce the correct equation. For example, if during the search of the unknown equation $f_3$ from Table 3 the user falsely assumes the result should contain the delta between $z_1$ and $z_3$, SciMED should find a more complex term of $f_3 = (13.08(z_1 - z_3)z_3 - (z_2 - z_3)z_3)/(z_2 z_4^2)$ instead of $f_3 = (13.08(z_1 - z_2)z_3)/(z_2 z_4^2)$. One can partially remedy this issue by introducing meta-learning to SciMED's pipeline[105,106]. Specifically, one can train an ML model on data of expert user's usage of SciMED that lead to positive results and generalize to similar tasks, thus, providing an initial recommendation for similar tasks[107,108]. Second, SciMED becomes increasingly sensitive to noise in the data the more complex the unknown equation becomes (as shown in Fig. 5). Thus, a more robust regularization method, inspired by recent accomplishments of ML and deep learning techniques, should be integrated to tackle this difficulty[109,110]. For example, one can extend the SITL approach to multiple users or even adopt it for different types of users[111]. Alternatively, one can test the performance of SciMED on more case examples to better understand how the performance declines in the presence of each type of noise, concluding ground rules of performance. Third, the GA-based and Las Vegas-based SR components are as robust as the elementary functions provided (see Table 4). For example, in the current case, SciMED would not be able to discover a symbolic expression with a square root of a variable unless it is given as an additional feature in the dataset. Hence, finding an optimal set of elementary functions for SciMED can be of great interest. Moreover, as stochastic functions are commonly used to describe natural phenomena, extending SciMED to support probabilistic equations will improve its usefulness. To add this capability, one can try a wide range of methods[112–114]. Similarly, differential equations are a common representation of physical models[115,116]. As SciMED is currently able to solve differential equations if and only if these are written as one-side differential equations (e.g., $du/dx = f(u, x)$), SciMED currently offers a limited solution as compared to the *SINDy* SR system[45]. This can be remedied in future work.

| Setting | Value |
|---|---|
| Number of samples | 10,000 or 400 |
| Test size | 20% or 75% |
| Number of times the AutoML & GA based SR components were run | 20 |
| GA based SR parsimony term range | 0.01-0.025 |
| Stability threshold for GA-based SR outcomes | 60% |
| T-test's $p$ value threshold for GA-based SR outcomes | 0.8 |
| Las vegas component size range | 5-17 |
| Number of syntactic data points ($\tau$) | 10,000 or 400 |
| The automatic machine learning sampling radius ($r$) | 7.5 |
| The automatic machine learning number of neighbors points ($\kappa$) | 3 |
| Termination criteria in hours | 12 |
| Levels of target noise (in all exp. but B) | 2% |
| Elementary functions used in the SR components | $add, sub, mul, div$ |

**Table 4.** Settings used in experiments A-E.

## Methods
The basic settings for experiments A-E are summarized in Table 4. The data for each experiment is generated in a table-like manner, demonstrated in Fig. 1; where columns represent variables with the last column being the target value calculated with them (e.g. $x_1, x_2, ..., f$ where $f(x_1, x_2, ...)$). The rows of the table contain the numbers representing samples of each variable. The functions used to generate the data of each experiment are listed in Table 3. Each of these functions is unknown to SciMED, and SciMED is required to deduce it from the data. The dataset for experiment D is generated similarly to the case of experiment C, except that the column containing the $z_4$ variable is deleted after the target column is generated, meaning there's no possible way of constructing the true equation for the target from the given variable columns.

All but experiment B had noise added to target values. The dataset of experiment A has 400 samples split 75/25% between training and testing. In the rest of the experiments, the dataset contains $10^4$ samples split 80/20% between training and testing. All GA and AutoML components of SciMED are tuned in a 5-fold manner. In experiment C, a grid search is performed on the parsimony term of the GA-based SR component within the range presented in Table 4. In all other experiments, a parsimony term of 0.02 is used instead of the grid search.

Additionally, we initiated SciMED, *AI Feynman* and *GP-GOMEA* on experiments A-E 20 times. In each experiment, the most repeated outcome was assigned as the result that the SR system found. These results (i.e., detected functions) are presented in Table 3.

## Data availibility
The datasets generated and analyzed with the open source Python library presented in this study are available in the SciMED repository, https://github.com/LironSimon/SciMED. The appendix is available in the supplementary materials.

### Author contributions
L.S.K.: Conceptualization, data curation, formal analysis, investigation, methodology, software, visualization, and writing–original draft. A.L.: Conceptualization, supervision, validation, and writing–review & editing. T.L.: Conceptualization, formal analysis, investigation, project administration, software, supervision, writing–original draft, and writing–review & editing.

### Funding
This research was partially supported by ISF (grant number 441/2) and Gordon Center for Energy Studies at Tel Aviv University.

### Competing interests
The authors declare no competing interests.

### Additional information
**Supplementary Information** The online version contains supplementary material available at https://doi.org/10.1038/s41598-023-28328-2.

**Correspondence** and requests for materials should be addressed to L.S.K.

**Reprints and permissions information** is available at www.nature.com/reprints.

**Publisher's note** Springer Nature remains neutral with regard to jurisdictional claims in published maps and institutional affiliations.